\def\year{2020}\relax
\documentclass[letterpaper]{article} 
\usepackage{aaai20}  
\usepackage{times}  
\usepackage{helvet} 
\usepackage{courier}  
\usepackage[hyphens]{url}  
\usepackage{graphicx} 
\usepackage{graphicx}  
\usepackage{CJK}
 
\usepackage{multirow}
\usepackage{float}
\usepackage{enumitem}

\usepackage{amsfonts}
\usepackage{amsmath}
\usepackage{setspace}
\usepackage[flushleft]{threeparttable}

\title{Chinese Named Entity Recognition Augmented with Lexicon Memory}
\author{Yi Zhou, Xiaoqing Zheng, Xuanjing Huang\\ 
School of Computer Science, Fudan University, Shanghai, China\\
\{yizhou17, zhengxq, xjhuang \}@fudan.edu.cn} 

\newcommand{\ours}{LEMON }
\newcommand{\oursp}{LEMON}
\newcommand{\formula}{ }
\newcommand{\reducebelow}{\setlength{\belowcaptionskip}{-15pt} }
\newcommand{\reducetable}{
    \setlength{\belowcaptionskip}{0pt} 
    \setlength{\abovecaptionskip}{0pt} 
}
\newcommand{\citet}[1]{\citeauthor{#1} \shortcite{#1}}

\begin{document}
\begin{CJK}{UTF8}{gbsn}

\maketitle

\begin{abstract}


Inspired by a concept of content-addressable retrieval from cognitive science, we propose a novel fragment-based model augmented with a lexicon-based memory for Chinese NER, in which both the character-level and word-level features are combined to generate better feature representations for possible name candidates. It is observed that locating the boundary information of entity names is useful in order to classify them into pre-defined categories. Position-dependent features, including prefix and suffix are introduced for NER in the form of distributed representation. The lexicon-based memory is used to help generate such position-dependent features and deal with the problem of out-of-vocabulary words. Experimental results showed that the proposed model, called LEMON, achieved state-of-the-art on four datasets. \footnote{https://github.com/dugu9sword/LEMON}

\end{abstract}

\section{Introduction}

Named Entity Recognition (NER) aims to locate and classify elements in sentences into pre-defined categories such as persons' names, organizations, locations, etc.
NER systems have been developed using linguistic rule-based techniques or statistical models. Rule-based systems identify names by applying linguistic grammar rules governing the derivation of names \cite{kapetanios:book2013}, while statistical models identify names based on the distribution of their components in a larger corpus \cite{lin:acl2009,nothman:ai2013}.
Recently, neural networks have been applied in NER \cite{collobert2011natural}, such as recurrent neural networks  \cite[RNNs]{huang2015bidirectional,lample2016neural} and encoder-decoder architectures \cite{shen2017grs,chen2018learning}. There are two reasons for the success of neural networks. On the one hand, neural network can memorize cases that have been seen after training. On the other hand,  they can be generalized to other unseen cases \cite{zhang2016understanding}. However, these models still suffer from two problems of ambiguous word boundaries and out-of-vocabulary words.

\textbf{Ambiguity of word boundaries:} Traditional approaches to Chinese NER can be divided into two paradigms: character-based and word-based models. Character-based models  are not effective enough due to lack of explicit word information \cite{he2008,liu2010chinese}, while word-based models suffer from the issue of error propagation, since word segmentation provides rather significant information for boundaries of named entities. \citet{zhang2018lattice} proposed a lattice-based model to encode a sequence of characters as well as every potential word that matches a lexicon \cite{li2018survey}. However, the important boundary features (prefix and suffix) for each name candidate might be blurred because they consider all possible segmentations, but only few of them are feasible, possibly introducing unnecessary noise. Named entities are often in the form of a \textit{fragment} (sequence of contiguous words) rather than a single \textit{character} or \textit{word} \cite{xu2017local}, which indicates that fragment-based models deserves further exploring.



\begin{figure}
\reducebelow

\begin{center}
\includegraphics[width=0.85\linewidth]{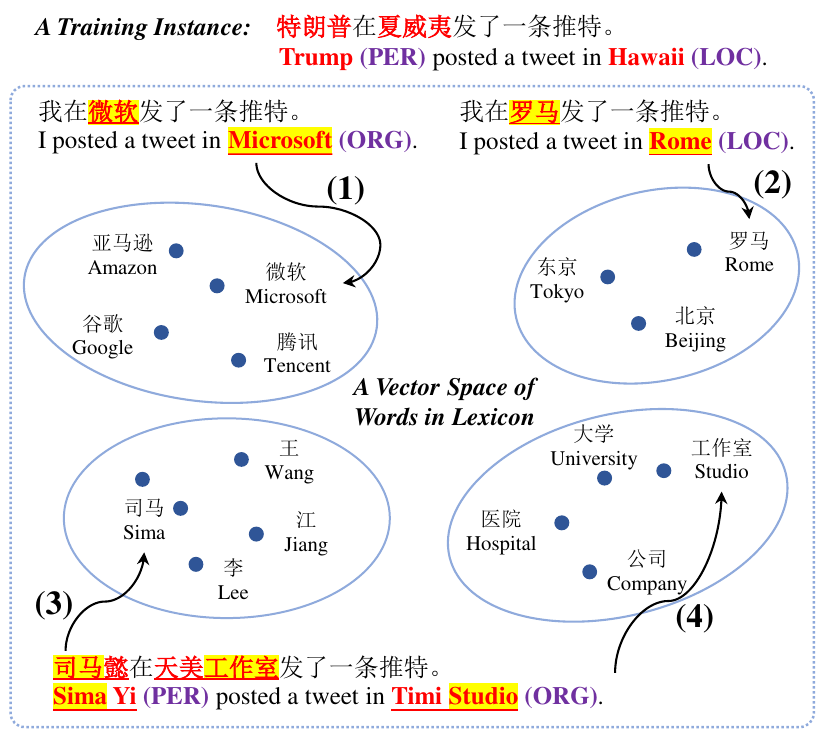}
\caption{Cases when position-dependent features benefit the identification of named entities. In cases (1) and (2), semantic information of the words is enough for recognition; while in cases (3) and (4), where the words may be out-of-vocabulary (e.g., names of persons or organizations not well known), syntactic information(prefix/suffix) will help.}
\label{fig:eg_oov}
\end{center}
\end{figure}

\textbf{Out-of-vocabulary words:} If word-level information can be harnessed in form of their embeddings, the adverse effect of unknown words could be much alleviated by leveraging a large unlabeled text corpus to learn word embeddings. As shown in Figure \ref{fig:eg_oov}, ``Microsoft'' would be classified with a higher probability into the correct category (namely organization) because its embedding is close to ``Google'', ``Amazon'', and so on in the embedding space. Such regular pattern can be also applied into location entities such as ``Rome'', ``Tokyo'' and ``Beijing''. However, for names of persons or organizations less well-known, such as ``司马懿'' (Sima Yi) and ``天美工作室'' (Timi Studio), which can not be found in the vocabulary, syntactic features may help. Most of the people's names start with a common Chinese surname followed by one or two characters. Organization names usually begin with the name of a city or country, and end with one of few words like ``公司 (company)'', ``大学 (university)'', ``医院 (hospital)'', etc. 

We propose a fragment-based approach to address the above problem, which combines information at different levels of granularity. \emph{Position-dependent} features, including prefix, suffix and infix deserve further investigation in the case of distributed representations. However not all fragments in a sentence are common words or phrases, we filter those rare ones with the assistance of a lexicon. It has proven fruitful to incorporate a lexicon (an external dictionary) for NER \cite{huang2015bidirectional,chiu2016named}, although such word-level features are added by string matching in a rigid, discrete manner. Constructing a lexicon via collecting information such as person surname list and geographical dictionary in a hand-crafted way is time-consuming, so it is worth exploring the possibility of deriving such features automatically from a large word corpus.

The fragment-based approach conforms to the way human recognizes names. Given a fragment, a person's attention will be drawn towards contents most relevant to her memory, which can be regarded as content-addressable retrieval, a concept borrowed from cognitive science to artificial intelligence  \cite{hassabis2017neuroscience}. From the viewpoint of cognitive systems, the biological brain does not learn by a single and global optimization principle \cite{hopfield1982neural}, but is modular and composed of distinct subsystems, such as memory and control which can interact with each other \cite{anderson2004integrated,shallice1988neuropsychology}.

Inspired by the findings from cognitive science, we propose a fragment-based model for Chinese NER augmented with a lexicon-based memory, called LEMON (LExicon-MemOry-augmented-Ner). The model consists of three submodules: a character encoder that imitates the process of scanning each character in an input sentence to grasp the global semantics, a fragment encoder that simulates the procedure of reading a sub-sequence (such as words or fragments) in a sentence, and a memory which stores massive words that have ever seen. A ranking algorithm is used to determine whether a fragment is a valid name and which category it belongs to by taking its prefix, suffix, and infix features into account. Experimental results showed that the proposed model achieved state-of-the-art results on four different benchmark datasets. 


\section{Related Work}

\subsection{Local Detection}
\citet{xu2017local} firstly presented a local detection approach for mention detection and name classification. Their model uses a fixed-size ordinary forgetting encoding (FOFE) to represent all fragments in the context \cite{zhang2015fixed}. Our model differs from theirs in that we adopts a character encoder to establish connections between the fragment and its context to provide global context features. Besides, position-dependent features are introduced for each candidate name via lexicon-based memory. 

\subsection{Attention Mechanism and Memory Network}
Attention mechanism was first proposed for machine translation  \cite{bahdanau2014neural,luong2015effective}, which learns an alignment between the source and target languages by estimating their correlation scores. It was also applied to NER in several ways: integrating character-level information by attending to characters \cite{rei2016attending}, capturing global context information by attending to different sentences in a document  \cite{xu2018improving}, and adopting an adaptive co-attention between texts and pictures \cite{zhang2018adaptive}. 
Memory networks was first introduced for question answering \cite{weston2014memory,hammerton2003named}, this study is among the first ones to incorporate word-level features by memory networks for NER.


\section{Model}

\begin{figure}
\reducebelow
\begin{center}
\includegraphics[width=0.8\linewidth]{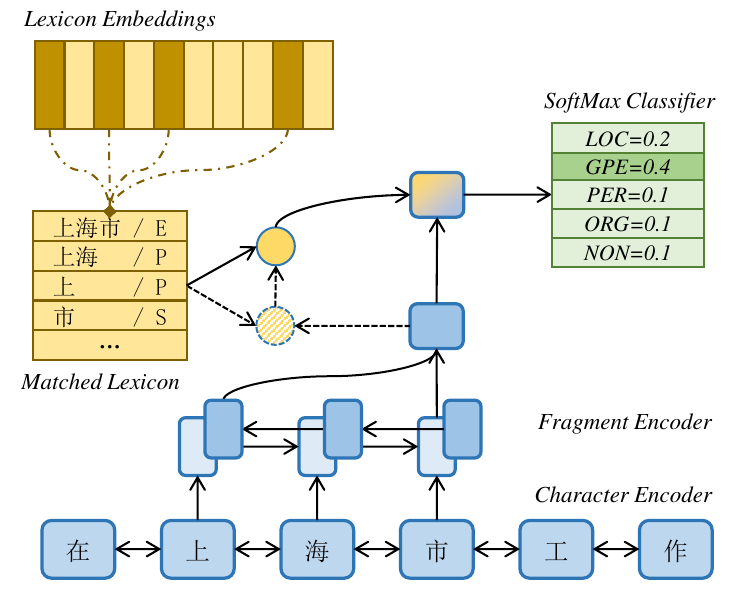}
\caption{The \ours model. Letter ``E'' denotes \textit{exactly} match, ``P'' \textit{prefix} match, and ``S'' \textit{suffix} match.}
\label{fig:model}
\end{center}
\end{figure}

We present  architecture of the proposed model in this section. 
As shown in Figure \ref{fig:model}, the \ours is mainly composed of three parts: a character encoder which maps each character into a its feature vector, a fragment encoder which encodes any variable-length sub-sequence in an input sentence into a fixed-sized vector representation, and a lexicon memory which is designed to help in disambiguating the word boundaries and dealing with the out-of-vocabulary problems by providing external syntactic and semantic features for possible words occurred in any fragment.

\subsection{Character Encoder} \label{char_encoder}

\begin{figure}
\reducebelow
\begin{center}
\includegraphics[width=0.8\linewidth]{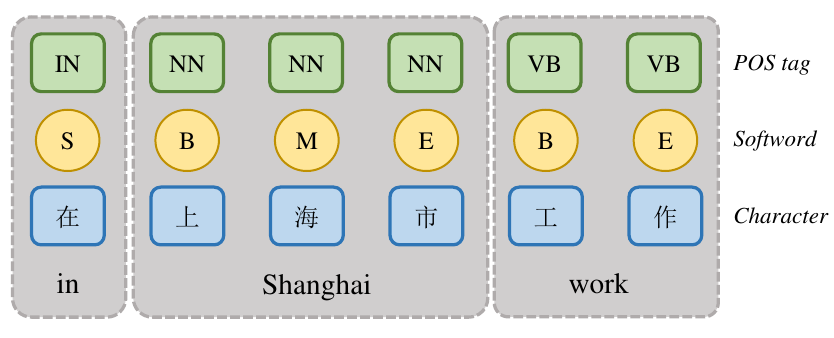}
\caption{Three different types of features used for each character. For word segmentation, each character will be assigned one of four possible boundary tags: ``B'' for a character located at the beginning of a word, ``M'' for that inside of a word, ``E'' for that at the end of a word, and ``S'' for a character that is a word by itself.}
\label{fig:character }
\end{center}
\end{figure}

Given a sentence $S=s_{[1:N]}$, each character $s_i$ is firstly mapped into its feature vector $w_i$. The information derived from the results of word segmentation and part-of-speech (POS) tagging has proven to be useful for NER tasks \cite{peng2016improving,yang2016combining}, and thus we augment the character representation with its soft-word and part-of-speech information. As shown in Figure \ref{fig:character }, the BMES scheme is used to represent the results of word segmentation \cite{xue2003chinese}. Each character is also assigned a POS tag as same as that of the word to which it belongs. The feature vector of each character is obtained by concatenating the feature vectors from the three parts as:
\begin{equation} \formula
    w_i = E^{char}_{char (i)} \oplus E^{seg}_{seg (i)} \oplus E^{pos}_{pos (i)}
\end{equation}
where $E^{char}$, $E^{seg}$, $E^{pos}$ are three look-up tables, $char (i)$, $seg (i)$, $pos (i)$ are indices of the characters, soft-word labels and POS tags.
The character encoder is used to get the context-aware representation of a character in a given sentence:
\begin{equation} \formula
    t_i = \textsc{Encoder}_{character } (w_{[1:N]})_i
\end{equation}
where $t_i \in \mathbb{R}^{d_t}$, $d_t$ is the dimensionality of the context-aware character representation. 
A few networks can be adopted as the character encoder, such as a bi-directional LSTM that is of great superiority in modelling long-distance dependencies \cite{elman1990finding,hochreiter1997long}, and a transformer that was firstly proposed for machine translation  \cite{vaswani2017attention} to capture the dependencies between different words with any distance in a sentence, which is gaining much attention recently.


\subsection{Fragment Encoder}
The fragment encoder is used to produce a feature vector for each $n$-gram in a sentence.
Given a sequence of characters, $T=t_{[i:j]} \in \mathbb{R}^{(j-i+1) \times d_t}$, where $d_t$
is the dimensionality of the
context-aware character representation, the fragment encoder learns to map the matrix $T$ to a fixed-sized vector $f_{[i:j]} \in \mathbb{R}^{d_f}$, where $d_f$ is the dimensionality of fragment embedding. 
\begin{equation} \formula
    f_{[i:j]} = \textsc{Encoder}_{fragment} (t_{[i:j]})
\end{equation}
where $[$$i$$:$$j$$]$ denotes a candidate fragment spanning from character $i$ to $j$.

Assuming that the maximum length of named entities is $m$, for a sentence consisting of $n$ characters, the number of all possible fragments would be $ \frac{m \times (2n-m+1)}{2} $. The complexity of enumerating all the fragments is $O(mn)$, which is rather time consuming. However, an inherent recursive structure helps to reduce the complexity, since the produced representations of shorter fragments can be used to generate those of longer ones, and all the fragments can be enumerated in $O(n)$ time. 

There are some methods that can be chosen as the fragment encoder. \citet{xu2017local} employs a Fixed-size Ordinary Forgetting Encoding (FOFE) as such encoder, which incorporates a forgetting factor to reflect position information \cite{zhang2015fixed}. The bag-of-words method that simply averages the representations of words or characters also can be used as a baseline encoder \cite{joulin2017bag,conneau2018you}.

\subsection{Lexicon Memory}

\subsubsection{Lexicon Construction}

The lexicon used in this study is not just a gazetteer (i.e., a vocabulary consisting of known named entities): it contains all the possible words extracted from a dataset, which allows us to leverage a large-scale unlabeled data to obtain rich features about words. Like \cite{zhang2018lattice}, the lexicon is obtained by automatically segmenting Chinese Giga-Word dataset\footnote{https://catalog.ldc.upenn.edu/LDC2011T13} and collecting the words. After that, the embeddings of the words in the lexicon are learned by word2vec \cite{mikolov2013distributed}. Due to the ambiguity of Chinese word segmentation, there may exist several smaller parts of a word in the lexicon, which reflects different levels of granularity. For example, the named entity ``财政部'' (Ministry of Finance), can exist along with ``财政'' (public finance), ``财'' (finance), ``政'' (administration), and ``部'' (ministry). The words at the finer level of granularity, such as ``部'' (ministry) can provide the finer word-formation features for NER.

\subsubsection{Lexicon Matching Modes} \label{mode}
Given a fragment $s_{[i:j]}$, we perform pattern matching on it over the constructed lexicon $V$. We define four types of matching modes as follows: 

\begin{itemize}[noitemsep]
    \item \textit{Exact matching}:
    If there exits one word in the lexicon that is exactly the same as the fragment, the word can be directly used to replace this fragment.
    \item \textit{k-prefix matching}: If the first $k$ characters of a fragment is matched with a word, we call it $k$-prefix matching. For example, a fragment ``$xyz$'' matches ``$xy$'' in a 2-\textit{prefix matching} mode. Such matching patterns provide informative features to identify the named entities whose prefixes are usually chosen from a limited number of words, such as commonly-used Chinese surnames like ``上官''  (Shangguan), and ``司马''  (Sima). 
    \item \textit{k-suffix matching}: We say a $k$-suffix matching if the last $k$ characters of a fragment, ``$xyz$'' is matched with a word ``$yz$''. Those matching patterns are quite useful to recognize the entities whose names end with one of few words. For example, many locations and organizations share the similar suffixes, such as ``省''  (Province), ``部''  (Ministry).
    \item \textit{Infix matching}: If a word can be found in the middle of a fragment, it is an infix matching. Its role is slightly different from the above modes, and such match serves as a hint that a fragment might contain a nested structure. 
\end{itemize}

Since the first (or last) one and two characters are relatively more important for NER, the results of different matching modes are grouped into multiple buckets according to their importance. We defined \oursp-$K$ in the way that for each distinct $k$, the feature derived from $k$-prefix (or $k$-suffix) matching is placed into a separate bucket if $k \leq K$, while the remaining features ($k > K$) are grouped into a single bucket. The value $K$ is a hyper-parameter.

\subsection{Attention over Lexicon Memory}

Memory networks provide us a feasible method to extract the relevant features from a lexicon-based memory with content-addressable retrieval \cite{weston2014memory,sukhbaatar2015end}.
Given a matching instance $m_{(l, c)}$, where $l$ denotes a matched word in the lexicon, and $c$ denotes one of the matching modes, they will be mapped into two feature vectors, and concatenated as a memory unit.
\begin{equation} \formula
    m_{(l, c)} = E^{lex}_l \oplus E^{mod}_c
\end{equation}
where $E^{lex} \in \mathbb{R}^{n_{lex} \times d_{lex}}$, $n_{lex}$ is the size of the lexicon, and $d_{lex}$ is the dimensionality of its vector space. $E^{mod} \in \mathbb{R}^{n_{mod} \times d_{mod} }$, $n_{mod}$ is the number of the matching modes, and $d_{mod}$ is the dimensionality of the feature vectors used to represent different matching modes.

For a fragment $s_{ [i:j]}$, we first find all its matched words, and then group them into multiple buckets as the way introduced in Section \ref{mode}, and finally assemble them into a matrix $M \in \mathbb{R}^{n_m \times d_m }$, which is a \textit{Lexicon Memory} dynamically built for the fragment, where $d_m = d_{lex}+d_{mod}$, and $n_m$ is the number of matching over the lexicon.

Given a fragment representation $f$ and its corresponding lexicon memory $M$, a scaled bi-linear attention is performed for $f$ over $M$ as follows \cite{luong2015effective,vaswani2017attention}:
\begin{equation} \formula
    attention (f, M) = softmax (\frac{f W M^\top}{\sqrt{d_m}})M
\end{equation}
where $W \in \mathbb{R}^{d_f \times d_m}$ is a learned parameter matrix. 

\subsection{Classification and Decoding}

\subsubsection{Training Objective}

For a fragment, its representation $f$ and the result of the attention over the lexicon $attention (f, M)$ are concatenated to produce the final representation $r_{ [i:j]}$. Such representation is then fed into a multi-layer feed-forward neural network to predict the labels of entities. If a fragment does not belong to any entity, it is labelled as ``NONE''. We choose to use a recently proposed focal loss as the training objective to mitigate the sample-imbalance problem~\cite{lin2017focal}.


\begin{equation} \formula
    loss ( p _ { t } ) 
    = 
    - \alpha_t   ( 1 - p_t ) ^ { \gamma } \log  ( p_t )
\end{equation}
where $p_t$ denotes the probability of the true label, $\alpha_t$ is a parameter vector for the true label which will be tuned during the training process, and $\gamma$ is a hyper-parameter that governs the relative importance of the positive samples with the negative ones. If all the values of $\alpha_t$, and $\gamma$ are set to $1$, the focal loss is reduced to the cross-entropy loss. 

\subsubsection{Decoding Strategy}
A decoding layer is stacked on top of the entity detector to resolve the issue that occasionally some overlapped fragments might be all recognized as valid entities \cite{xu2017local}:
\begin{itemize}[noitemsep] 
    \item A threshold $\rho$ is used to filter the results. A fragment is identified as an entity if the model assigns the highest probability to this entity type and the probability is greater than $\rho$; otherwise it will be recognized as ``NONE''.
    \item If a recognized entity contains another  candidate (nested) entity, only the outer entity will be remained for the further processing.
    \item If two identified entities overlap each other, only the one with higher probability is kept.
\end{itemize}
We found that such decoding strategy works well although it runs in a greedy way. This strategy also can be used to recognize nested entities just by removing the second step.

\begin{table}[ht]
	\begin{center}
		\small
		\caption{Statistics of datasets}\label{table:statistics}
		\begin{tabular}{c|c|c|c|c}
			\hline
			\hline
			\textbf{Dataset} & \textbf{\#Train}  & \textbf{\#Dev} & \textbf{\#Test} & \textbf{Domain} \\ 
			\hline
			\multicolumn{1}{l|}{OntoNotes-4} & 15.7k & 4.3k & 4.3k & News\\
			\multicolumn{1}{l|}{MSRA} & 46.4k & - & 4.4k & News\\
			\multicolumn{1}{l|}{Weibo} & 1.4k & 0.27k & 0.27k & Social Media\\
			\multicolumn{1}{l|}{Resume} & 3.8k & 0.46k & 0.48k & Resume\\
			\hline
			\hline
		\end{tabular}
	\end{center}
\end{table}


\begin{table*}[ht]
\reducetable

\begin{center}
\caption{Results on OntoNotes-4 development set with different model architectures. }\label{table:model-arch}
\resizebox{\textwidth}{!}{
    \begin{tabular}{ccccc|ccc|ccc}
        \hline
        \hline
       & &  \multicolumn{1}{|c|}{\textbf{P (\%)}} & \multicolumn{1}{c|}{\textbf{R (\%)}} & \multicolumn{1}{c|}{\textbf{F1 (\%)}}
         & \multicolumn{1}{c|}{\textbf{P (\%)}} & \multicolumn{1}{c|}{\textbf{R (\%)}} & \multicolumn{1}{c|}{\textbf{F1 (\%)}} 
         & \multicolumn{1}{c|}{\textbf{P (\%)}} & \multicolumn{1}{c|}{\textbf{R (\%)}} & \multicolumn{1}{c}{\textbf{F1 (\%)}} \\ 
        \cline{3-11}
        \multicolumn{2}{c|}{\textbf{fragment \textbackslash{} Character }} 
        & \multicolumn{3}{c|}{\textbf{Baseline}} 
        & \multicolumn{3}{c|}{\textbf{Transformer}} 
        & \multicolumn{3}{c}{\textbf{Bi-RNN}} \\ 
         \hline
        \multirow{6}{*}{\rotatebox{90}{\textbf{Gold}}} 
       &  \multicolumn{1}{|l|}{ \textbf{BOW} } 
        & 72.40 & 62.03 & 66.81 & - & - & - & 73.60 & 69.08 & 71.27 \\
       &  \multicolumn{1}{|l|}{ \textbf{FOFE} } 
        & 75.52 & 64.86 & 69.78 & 64.35 & 54.04 & 58.74 & 76.93 & 70.43 & 73.54 \\
       &  \multicolumn{1}{|l|}{ \textbf{Bi-RNN} } 
        & 73.68 & 69.74 & 71.66 & 59.92 & 54.87 & 57.28 & 71.51 & 73.66 & 72.57  \\\cline{2-11}
        
       &  \multicolumn{1}{|l|}{ \textbf{BOW + Lex} } 
        & 78.77 & 70.40 & 74.35 (+7.54) & 76.73 & 73.48 & 75.07 (+8.26) & 78.27 & 75.34 & 76.78 (+5.51) \\
       &  \multicolumn{1}{|l|}{ \textbf{FOFE + Lex} } 
        & 77.33 & 71.90 & 74.52 (+4.74) & 79.92 & 72.65 & 76.11 (+17.37) & 79.49 & 73.77 & 76.53 (+2.99) \\
       &  \multicolumn{1}{|l|}{ \textbf{Bi-RNN + Lex} } 
        & 77.40 & 74.39 & 75.87 (+4.21) & 79.62 & 73.87 & 76.64  (+19.36) & 81.12 & 75.18 & \textbf{78.04 (+5.47)}  \\
        
        \hline
        \multirow{6}{*}{\rotatebox{90}{\textbf{Auto}}} 
       &  \multicolumn{1}{|l|}{ \textbf{BOW} } 
        & 76.67 & 56.24 & 64.88 & - & - & - & 75.39 & 61.36 & 66.92 \\
       &  \multicolumn{1}{|l|}{ \textbf{FOFE} } 
        & 71.66 & 58.21 & 64.24 & 73.17 & 61.75 & 66.98 & 76.20 & 61.65 & 68.16 \\
       &  \multicolumn{1}{|l|}{ \textbf{Bi-RNN} } 
        & 74.60 & 63.67 & 68.70 & 72.05 & 63.52 & 67.52 & 76.73 & 63.70 & 69.61  \\\cline{2-11}
        
       &  \multicolumn{1}{|l|}{ \textbf{BOW + Lex} } 
        & 76.33 & 64.75 & 70.06 (+5.18) & 73.96 & 64.69 & 69.02 (+4.14) & 78.42 & 67.06 & 72.30 (+5.38) \\
       &  \multicolumn{1}{|l|}{ \textbf{FOFE + Lex} } 
        & 77.24 & 63.91 & 69.95 (+5.71) & 78.46 & 62.93 & 69.85 (+2.87) & 76.24 & 68.76 & 72.31 (+4.15) \\
       &  \multicolumn{1}{|l|}{ \textbf{Bi-RNN + Lex} } 
        & 77.62 & 66.32 & 71.53 (+2.83) & 76.79 & 67.09 & 71.61 (+4.09) & 76.57 & 69.54 & \textbf{72.89 (+3.28)} \\
       \hline
       \hline
    \end{tabular}
}
    \begin{tablenotes} \small
    \item The heading with a word ``gold'' denotes that gold segmentation and part-of-speech tags are used, while ``auto'' denotes that they are automatically generated by the THULAC toolkit.
    \end{tablenotes}
\end{center}
\end{table*}

\begin{table*}[ht]\small

\reducetable

\begin{center}
\caption{Results on the OntoNotes-4 development set with different features}\label{table:features}
\setlength{\tabcolsep}{5.1mm}{
\begin{tabular}{ccccc|ccc}
    \hline
    \hline
   & & \multicolumn{1}{|c|}{\textbf{P (\%)}} & \multicolumn{1}{c|}{\textbf{R (\%)}} & \multicolumn{1}{c|}{\textbf{F1 (\%)}}
     & \multicolumn{1}{c|}{\textbf{P (\%)}} & \multicolumn{1}{c|}{\textbf{R (\%)}} & \multicolumn{1}{c}{\textbf{F1 (\%)}}  \\ 
    \cline{3-8}
   \multicolumn{2}{c|}{\textbf{Features \textbackslash{} Data}}  & \multicolumn{3}{c|}{\textbf{Ground truth}} 
     & \multicolumn{3}{c}{\textbf{Automatically labelled}}\\ 
     \hline
    \multirow{4}{*}{\rotatebox{90}{\textbf{NCRF}}} 
   &  \multicolumn{1}{|l|}{ \textbf{char} } & 66.37 & 60.21 & 63.14 & - & - & -  \\
   &  \multicolumn{1}{|l|}{ \textbf{char + seg} } & 70.58 & 69.96 & 70.27 & 70.77 & 63.33 & 66.85 \\           
   &  \multicolumn{1}{|l|}{ \textbf{char + pos} } & 71.81 & 74.48 & 73.12 & 70.20 & \textbf{70.26} & 70.23 \\
   &  \multicolumn{1}{|l|}{ \textbf{char + seg + pos} } & 75.63 & 72.35 & 73.08 & 72.88 & 68.18 & 70.45  \\
    \hline
    \multirow{4}{*}{\rotatebox{90}{\textbf{No Lex}}} 
   &  \multicolumn{1}{|l|}{ \textbf{char} } & 67.6 & 55.03 & 60.67 & - & - & - \\
   &  \multicolumn{1}{|l|}{ \textbf{char + seg} } & 72.16 & 66.09 & 68.99 & 70.48 & 62.65 & 66.33 \\            
   &  \multicolumn{1}{|l|}{ \textbf{char + pos} } & 74.39 & 65.44 & 69.63 & 72.87 & 63.73 & 67.99 \\
   &  \multicolumn{1}{|l|}{ \textbf{char + seg + pos} } & 74.97 & 72.23 & 73.58 & 76.29 & 64.43 & 69.86   \\
   \hline
   \multirow{4}{*}{\rotatebox{90}{\textbf{Lex}}} 
   &  \multicolumn{1}{|l|}{ \textbf{char} } & 77.27 & 60.73 & 68.01 & - & - & -  \\
   &  \multicolumn{1}{|l|}{ \textbf{char + seg} } & 78.40 & 70.75 & 74.38 & 76.11 & 64.63 & 69.91 \\            
   &  \multicolumn{1}{|l|}{ \textbf{char + pos} } & 77.71 & 72.35 & 74.93 & 77.46 & 66.89 & 71.79 \\
   &  \multicolumn{1}{|l|}{ \textbf{char + seg + pos} } & \textbf{78.70} & \textbf{74.95} & \textbf{76.78} & \textbf{76.41} & 68.61 & \textbf{72.30}   \\
    \hline
    \hline
\end{tabular}
}
\end{center}
\end{table*}

\section{Experiments}

\subsection{Experiments Settings}

\subsubsection{Datasets}
We evaluated our model on four different datasets: OntoNotes-4  \cite{weischedel2011ontonotes}, MSRA  \cite{levow2006third}, Weibo NER \cite{peng2016improving,peng2015named}, and Resume \cite{zhang2018lattice} datasets. The statistics of those four datasets are given in Table \ref{table:statistics}. As mentioned in Section \ref{char_encoder}, each character needs to be assigned with a soft-word label as well as a POS tag. All the datasets were segmented and tagged by using THULAC toolkit \cite{sun2016thulac}, which achieved about $88\%$ of F1-score in the word segmentation on the datasets. For OntoNotes-4 dataset, the gold segmentation and part-of-speech tags are available, and we reported the NER results both with and without gold segmentation and POS tags.

\subsubsection{Training Details}
The proposed model is implemented by PyTorch deep learning framework \cite{paszke2017automatic}. 
We pretrained the word embeddings and character embeddings on  Chinese Giga-word by Word2vec. We tuned all the hyper-parameters on the development set of OntoNotes-4 dataset. The dimensionalities of word embeddings and character embeddings were all set to $50$, and the dimensionalities of soft-word embeddings and POS tag embeddings were both set to $25$. Dropout mechanism was applied to the character encoder at the embedding layer with a drop rate of $0.3$. All learned parameters are updated by the Adam Optimizer \cite{kingma2014adam}. It is worth mentioning that we use a sparse version of the Adam Optimizer\footnote{https://pytorch.org/docs/stable/optim.html\#torch.optim. SparseAdam} to update the learned embedding parameters. The learning rate was set to $1e-3$, and the weight decay to $1e-7$\footnote{The weight decay was set to $1e-5$ for the Weibo NER dataset, otherwise the network will be hard to converge.}.


\subsection{Experiments on OntoNotes-4}

We carried out a set of preliminary experiments on the development set of OntoNotes-4 to optimize the architecture by trying few different components, and to gain some understanding of how the choice of features impacts upon the performance.

\subsubsection{Evaluation with Different Architectures}
We tried several combinations of different character and fragment encoders to find a suitable configuration for NER. Three different types of networks were tested as the character encoder, and we also tried three different architectures for the fragment encoder. An embedding look-up layer serves as a baseline for the character encoder. Besides, two popular sequence models including a transformer (6-layer, 8-heads, 512-dim) and a Bi-directional LSTM (2-layer, 256-dim) are compared. As to the fragment encoder, we conducted experiments with Bag-of-Words (BOW), FOFE ($\alpha=0.5$) and bi-directional LSTM. \citet{xu2017local} used an embedding look-up layer as the character encoder and a FOFE as the fragment encoder, and they predict the type of a candidate $n$-gram with the help of its left and right contexts. We tried to integrate such context information for NER as \citet{xu2017local}, but the results of preliminary experiments showed that its contribution in performance is negligible.

The results of different combinations on the development set of OntoNotes-4 are shown in Table \ref{table:model-arch}. The performances of all models will decrease of approximately $4\%$ in F1-score if we used the results of word segmentation and POS-tagging automatically generated by THULAC toolkit instead of the ground truth. It shows that the NER performance is significantly influenced by the results of the upstream tasks through the error propagation.

\textbf{Character Encoder:} Bi-RNN always outperforms other character encoders due to its ability in modelling long-term dependencies. The transformer contributes a little, and performs slightly better than the baseline although it achieved a great success in the machine translation. One reasonable explanation is that the number of training sentences is not sufficient enough to fit the model capacity of the transformer \cite{devlin2018bert}. 

\textbf{Fragment Encoder:} Bi-RNN surpasses other encoders, especially when the character encoder is not built based on the Bi-RNN. BOW performs inferior to others since it is unable to model the order information of a sequence, which is critical for the entity recognition. FOFE learns to produce a linear combination of the representations of words in a sub-sequence, which is less flexible than the Bi-RNN in the sequence modeling since the latter is capable of learning non-linear combinations.

\textbf{Lexicon Memory:} The incorporation of lexicon memory greatly boosts the results of any combination of components, with an average increase of about $5\%$ in F1-score. It can be taken as a strong evidence that the introduced lexicon memory can enhance the model's performance in NER.

\subsubsection{Feature Combinations}
The significance of different features is shown in Table \ref{table:features}. We also trained a LSTM-CRF model as a traditional approach for comparison by NCRF++, an open source neural sequence labeling toolkit \cite{yang2018ncrf}. The experimental results demonstrate that the features derived from the word segmentation and POS-tagging always benefit to all the models no matter they are labeled by human or produced by automatic toolkit.


The \ours still beats the LSTM-CRF-based model by $5\%$ in F1-score without using any word segmentation or part-of-speech information, which shows that the introduced lexicon memory provides the valuable position-dependent and word-level features via the attention mechanism.

\subsection{Results}

\reducetable

\begin{table}[ht]
	\begin{center}
		\small
		\caption{Results on the MSRA dataset}\label{table:msra-result}
		\begin{tabular}{c|c|c|c}
			\hline
			\hline
			\textbf{Model} & \textbf{P (\%)} & \textbf{R (\%)}  & \textbf{F1 (\%)} \\ 
			\hline
			\multicolumn{1}{l|}{ \cite{chen2006chinese}} & 91.22 & 81.71 & 86.20 \\
			\multicolumn{1}{l|}{ \cite{zhang2006sighan}} & 92.20 & 90.08 & 91.18 \\
			\multicolumn{1}{l|}{ \cite{lu2016multi}} & - & - & 87.94 \\
			\multicolumn{1}{l|}{ \cite{dong2016character}} & 91.28 & 90.62 & 90.95 \\
			\multicolumn{1}{l|}{ \cite{zhang2018lattice}} & 93.57 & \textbf{92.79} & 93.18 \\ 
            \multicolumn{1}{l|}{\ours}  & \textbf{95.39} & 91.77 &  \textbf{93.55} \\
			\hline
			\hline
		\end{tabular}
	\end{center}
\end{table}

\begin{table}[ht]

\reducetable

	\begin{center}
		\small
		\caption{Results on the Resume NER dataset}\label{table:resume-result}
		\begin{tabular}{c|c|c|c}
			\hline
			\hline
			\textbf{Model} & \textbf{P (\%)} & \textbf{R (\%)}  & \textbf{F1 (\%)} \\ 
			\hline
			\multicolumn{1}{l|}{word$\dag$} & 93.72 & 93.44 & 93.58 \\ 
            \multicolumn{1}{l|}{word+char+bichar$\dag$} & 94.07 & 94.42 & 94.24 \\ 
			\multicolumn{1}{l|}{char $\dag$} & 93.66 & 93.31 & 93.48 \\ 
			\multicolumn{1}{l|}{char+bichar+softword$\dag$} & 94.53 & \textbf{94.29} & 94.41 \\ 
			\multicolumn{1}{l|}{ \cite{zhang2018lattice}} & 94.81 & 94.11 & 94.46 \\ 
            \multicolumn{1}{l|}{\ours} & \textbf{95.59} & 94.07  & \textbf{94.82} \\
			\hline
			\hline
		\end{tabular}
		\begin{tablenotes}
            \small
            Models indicated with $\dag$ are those in which the sequence labeling technique is used with LSTM + CRF. Results are extracted from  \cite{zhang2018lattice}.
        \end{tablenotes}
	\end{center}
\end{table}

\begin{table}[ht]

\reducetable

	\begin{center}
		\small
		\caption{Results on the OntoNotes-4 dataset}\label{table:onto-result}
		\begin{tabular}{c|c|c|c}
			\hline
			\hline
			\textbf{Model} & \textbf{P (\%)} & \textbf{R (\%)}  & \textbf{F1 (\%)} \\ 
			\hline
			\multicolumn{1}{l|}{ \cite{wang2013effective} $\dag$} & 76.43 & 72.32 & 74.32\\
			\multicolumn{1}{l|}{ \cite{che2013named} $\dag$} & 77.71 & 72.51 & 75.02\\
			\multicolumn{1}{l|}{ \cite{yang2016combining} $\dag$} & 72.98 & \textbf{80.15} & 76.40 \\
			\multicolumn{1}{l|}{\ours $\dag$} & 79.27 & 78.29 & \textbf{78.78}\\
			\hline 
			\multicolumn{1}{l|}{ \cite{zhang2018lattice}} & 76.35 & 71.56 & 73.88 \\ 
            \multicolumn{1}{l|}{\ours} & \textbf{80.61} & 71.05 & \textbf{75.53} \\
			\hline
			\hline
		\end{tabular}
		\begin{tablenotes}
            \small
            The model indicated with $\dag$ denotes that gold word segmentation are used.
        \end{tablenotes}
	\end{center}
\end{table}

\begin{table}[ht]

\reducetable

	\begin{center}
		\small
		\caption{Results on the Weibo NER dataset}\label{table:weibo-result}
		\begin{tabular}{c|c|c|c}
			\hline
			\hline
			\textbf{Model} & \textbf{P (\%)} & \textbf{R (\%)}  & \textbf{F1 (\%)} \\ 
			\hline
			\multicolumn{1}{l|}{ \cite{peng2016improving}} & - & - & 58.99 \\
			\multicolumn{1}{l|}{ \cite{he2017unified}} & - & - & 58.23 \\
			\multicolumn{1}{l|}{ \cite{zhang2018lattice}} & - & - & 58.79 \\ 

            \multicolumn{1}{l|}{\ours} & 70.86 & 55.42  & \textbf{62.19} \\
			\hline
			\hline
		\end{tabular}
	\end{center}
\end{table}

The \oursp-2 achieved state-of-the-art results on all the four datasets. As shown in Table \ref{table:msra-result} and \ref{table:resume-result}, the \ours performs slightly better than the Lattice LSTM on the MSRA and Resume NER Datasets. Our model also achieved the highest F1-score on the OntoNotes-4 (see Table \ref{table:onto-result}). Note that the Weibo NER data is extracted from the social media, it is full of non-standard expressions and only contains about $1.4$k samples. The problems of out-of-vocabulary words and ambiguity of word boundaries become more serious for NER on this dataset. However, the \ours still outperforms other models with a fairly significant margin (at least $3\%$ increase), as we can see in Table \ref{table:weibo-result}.


\begin{figure*}[ht]
\centering
\begin{minipage}[t]{0.68\textwidth}
\includegraphics[width=\linewidth]{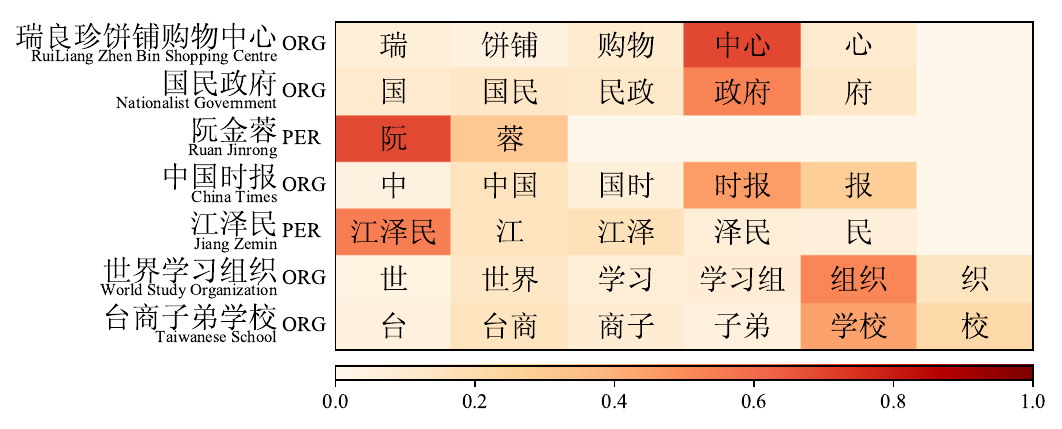}
\caption{An example heat map. Note that the attentions are sharp for \\
those words particularly useful for NER.}
\label{fig:attention}
\end{minipage}
\begin{minipage}[t]{0.28\textwidth}
\includegraphics[width=\linewidth]{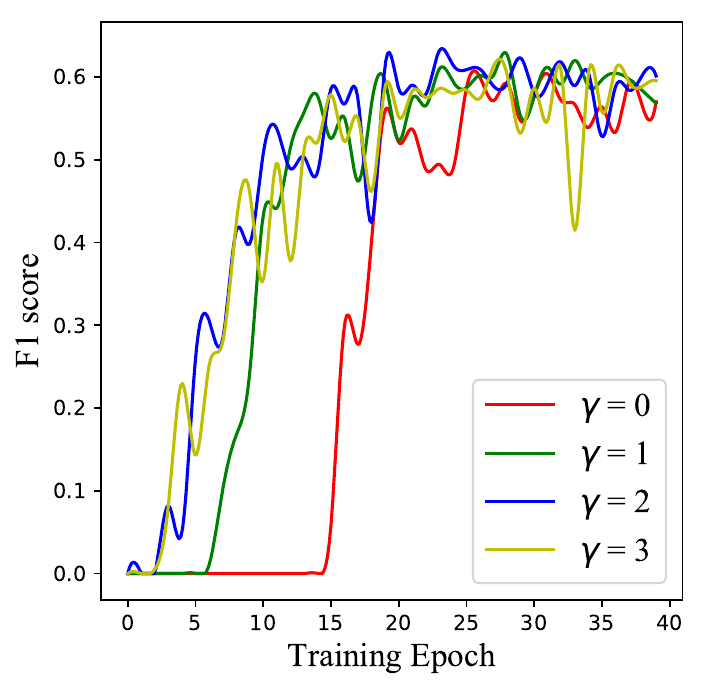}
\caption{F1-score versus training epochs.}
\label{fig:focal}
\end{minipage}
\end{figure*}

\begin{figure*}[ht]
\reducebelow
\begin{center}
\includegraphics[width=0.9\linewidth]{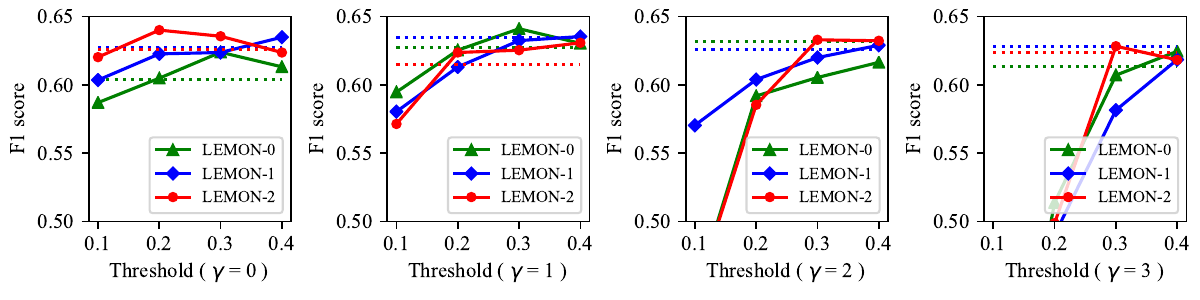}
\caption{Best F1-score with regards to decoding thresholds under different values of gamma, the horizontal line represents the results without the decoding step.}
\label{fig:threshold}
\end{center}
\end{figure*}


\subsection{Discussion}

We conducted experiments on the Weibo NER dataset to study the influence of attention over lexicon memory, and how the choice of values of the thresholds and focal loss coefficients impact upon the performance.

\subsubsection{Attention over Lexicon Memory}

Figure \ref{fig:attention} illustrates which words will be given more weights computed by the attention operations over the lexicon memory. As shown in the heat map, the model can learn to assign more weights on the key words of named entities, and the attentions are sharp for those words particularly informative for NER. 

Taking the entities of ``ORG'' (organization) as examples, more weights are placed to the last two characters, such as ``中心'' (center), ``政府'' (government), ``学校'' (school), ``组织'' (organization), etc. It is in accordance with the common sense that the last characters are more important in identifying Chinese names of organizations. We also found the similar phenomenon when recognizing person names. For instance, famous names such as ``江泽民'' (Zemin Jiang) can be matched exactly and recognized as a person name, while for names of less well-known persons, the first character (i.e. surname) tends to be given more attention.

\subsubsection{Decoding Threshold Settings}

We reported the F1-scores for different settings of \ours on the development set of Weibo NER in Figure \ref{fig:threshold}. \oursp-2 generally performs better than \oursp-0 and \oursp-1, since the features derived from $1$- and $2$-prefix and suffix matching are useful for NER and they cannot be mixed into a single bucket as we described in Section \ref{mode}.

We found that the best value of the threshold $\rho$ is in range of $0.2$ and $0.3$. As shown in Figure \ref{fig:threshold}, if the value of $\gamma$ is greater than or equal to $2$, the performances are more sensitive to the values of threshold $\rho$. The performance will drop dramatically if $\rho$ is less than $0.3$ and $\gamma$ is set to $3$. One possible explanation is that the focal loss tends to update the parameters by a far larger step for the samples that are hard to be recognized, especially when the probabilities assigned for those samples are pretty low. 


\subsubsection{Coefficients of Focal Loss}

We compared the speed of convergence versus  different values of $\gamma$ used in focal loss in Figure \ref{fig:focal}. If $\gamma$ is set to zero, the focal loss will be reduced to the cross entropy loss. When the cross entropy loss is used, the model is trapped at an extreme low performance for $15$ epochs, which indicates that such loss is not optimal for the situation with severe sample imbalance. Note that the models usually suffer from the problem of sample imbalance in NER because most candidates will be labelled as ``NONE''. The model with the focal loss converges relatively faster because this loss will adaptively assign different update steps to mis-classified samples according to how hard they are recognized. Although the model trained with the focal loss did not outperform that with the cross entropy, it does help to speed up the training process.

\section{Conclusion}

Observing that Chinese names are usually formed in some distinct patterns and the features derived from their prefix and suffix are particularly useful to identify them, a fragment-based model augmented with position-dependent features learned from a lexicon is introduced for Chinese NER tasks. Experimental results showed that the model using position-dependent features and lexicon-based memory achieved state-of-the-art on four different NER datasets.






\small{
\bibliographystyle{aaai}
\bibliography{aaai}
}

\end{CJK}

\end{document}